\def\year{2022}\relax
\documentclass[letterpaper]{article} 
\usepackage{aaai22}  
\usepackage{times}  
\usepackage{helvet}  
\usepackage{courier}  
\usepackage{hyperref}
\usepackage{graphicx} 
\usepackage{natbib}  
\usepackage{caption} 
\DeclareCaptionStyle{ruled}{labelfont=normalfont,labelsep=colon,strut=off} 
\usepackage{algorithm}
\usepackage{algorithmic}

\usepackage{newfloat}
\usepackage{listings}
\floatstyle{ruled}
\newfloat{listing}{tb}{lst}{}
\floatname{listing}{Listing}
\title{Godot Reinforcement Learning Agents}
\author{
    Edward Beeching,\textsuperscript{\rm 1}
    Jilles Debangoye, \textsuperscript{\rm 1}
    Olivier Simonin \textsuperscript{\rm 1}
    Christian Wolf\textsuperscript{\rm 2}
}
\affiliations{
    \textsuperscript{\rm 1}INRIA Chroma team, CITI Laboratory. INSA-Lyon, France.
\\
    \textsuperscript{\rm 2}Université de Lyon, INSA-Lyon, LIRIS, CNRS, France.\\
    Project page \url{https://edbeeching.github.io/papers/gdrl}
   
}

\begin{document}
\maketitle
\begin{abstract}
We present Godot Reinforcement Learning (RL) Agents, an open-source interface for developing environments and agents in the Godot Game Engine. The Godot RL Agents interface allows the design, creation and learning of agent behaviors in challenging 2D and 3D environments with various on-policy and off-policy Deep RL algorithms. We provide a standard Gym interface, with wrappers for learning in the Ray RLlib and Stable Baselines RL frameworks. This allows users access to over 20 state of the art on-policy, off-policy and multi-agent RL algorithms. The framework is a versatile tool that allows researchers and game designers the ability to create environments with discrete, continuous and mixed action spaces. The interface is relatively performant, with 12k interactions per second on a high end laptop computer, when parallized on 4 CPU cores. 
An overview video is available here: \url{https://youtu.be/g1MlZSFqIj4}.

\end{abstract}
\section{Introduction}
Over the next decade advances in AI algorithms, notably in the fields of Deep Learning \cite{yan2015deep} and Deep Reinforcement Learning, are primed to revolutionize the Video Games industry. Customizable enemies, worlds and story telling will lead to diverse game-play experiences and new genres of games. Currently the field is dominated by large organizations and pay to use engines that have the resources to create such AI enhanced agents. The objective of the Godot RL Agents package is to lower the bar of accessibility; so that game developers, researchers and hobbyists can take their idea from creation to publication end-to-end with an open-source and free package. The Godot RL Agent provides wrappers for two well known open-source Deep Reinforcement Learning libraries: Ray Rllib \cite{liang2018rllib} and Stable Baselines \cite{stable-baselines}. In addition the interface is compatible with any RL framework that can interact with a gym wrapper. Although the Godot RL Agents interface was predominantly designed  for the creation and prototyping of RL approaches in video games, there is nothing prohibiting users from the implementation of tasks involving mobile robotics, grasping or designing complex memory-based scenarios to test the limitations of current Deep RL architectures and algorithms.
\noindent The contributions the Godot RL Agents packages are as follows:
\begin{itemize}
    \item A free and open source tool for Deep RL research and game development.
    \item By providing an interface between the Godot Game Engine and Deep RL algorithms, we enable game creators to imbue their non-player characters with unique behaviors, learned through interaction. 
    \item The framework enables automated game-play testing with an RL agent.
    \item The Godot RL Agents framework enables researchers and game designers the flexibility to design, create and perform rapid iteration on new ideas and scenarios.
\end{itemize}

\noindent The library is open-source with an MIT licence, code is available on our github page:\\ \url{https://github.com/edbeeching/godot\_rl\_agents}\\

\begin{figure}
    \centering
    \includegraphics[width=0.50\textwidth]{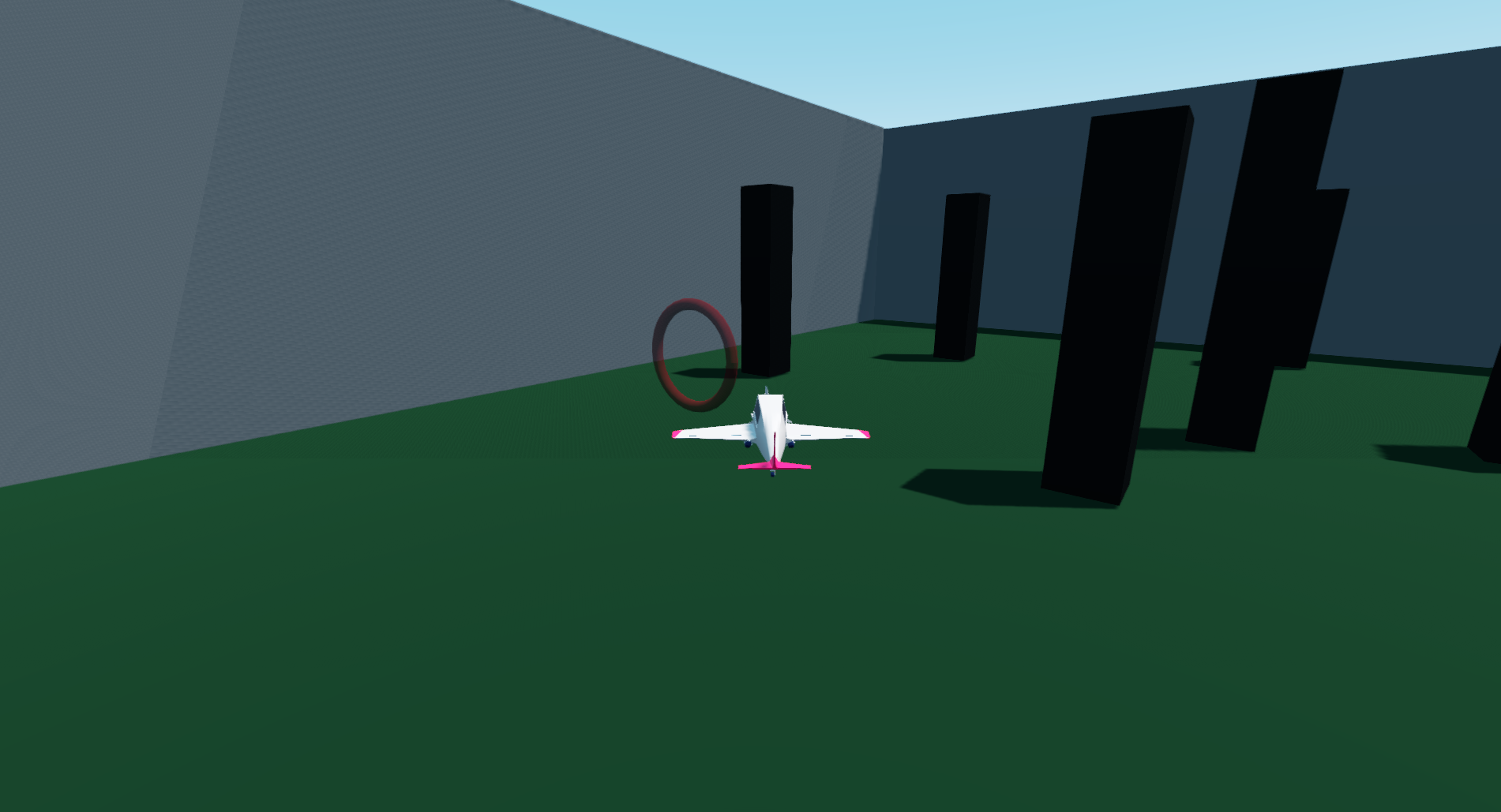}
    \caption{An example Godot RL Agents environment \textit{Fly By}, where an agent learns to perform way-point based aerial navigation in a 3D space.}
    \label{fig:random_method_coverage}
\end{figure}
\begin{figure*}[t]
    \centering
    \includegraphics[width=0.8\textwidth]{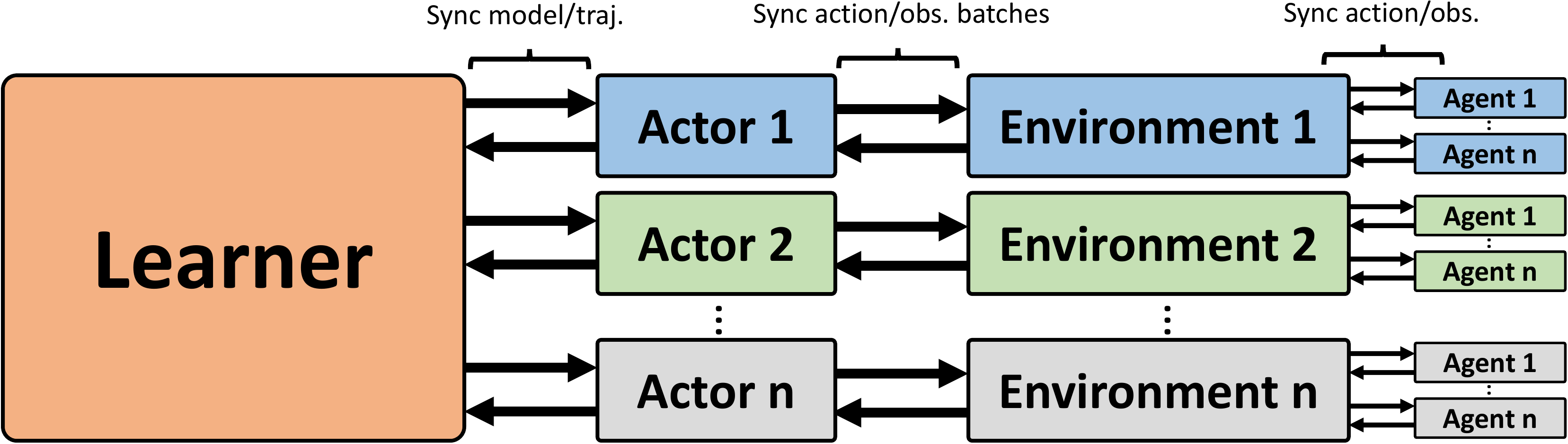}
    \caption{The training architecture used for the Godot RL Agents interface. Parallel actor processes collect trajectories from several environment executables. Each environment contains multiple parallel agents, each with their own instantiation of the environment.}
    \label{fig:learning_architecture}
\end{figure*}
\section{Related work}\label{sec:related_work}
The last decade has seen rapid advancement in the field of artificial intelligence, this has been driven by the application of Deep Learning approaches and in particular Convolutional Neural Networks \cite{fukushima1982neocognitron, Lecun1998Gradient-BasedMethod} . Supervised Deep Learning in its current state often considered to be narrow AI, that performs particularly well at performing predictions on static images but does not \textbf{learn from interaction}. 

\subsection{Deep Reinforcement Learning}
Reinforcement Learning approaches provide the ability to learn in sequential decision making problems, where the objective is to maximize accumulated reward. As we encounter large state spaces, continuous actions, partial observability \cite{aastrom1965optimal} and the desire to generalize to unseen environment configurations; Deep Reinforcement Learning (RL) provides a general framework for solving these problems. In recent years the field of Deep RL has gained attention with successes on board games  \cite{SilverMasteringSearch} and Atari Games \cite{Mnih2015Human-levelLearning}. One key component was the application of deep neural networks to frames from the environment or game board states. Recent works that have applied Deep RL for the control of an agent in 3D environments such as maze navigation are \cite{Mirowski2016LearningEnvironments} and \cite{Jaderberg2016ReinforcementTasks} which explored the use of auxiliary tasks such as depth prediction, loop detection and reward prediction to accelerate learning. Meta RL approaches for 3D navigation have been applied by \cite{Wang2016LearningLearn} and \cite{Lample} also accelerated the learning process in 3D environments by prediction of tailored game features. There has also been recent work in the use of street-view scenes in order train an agent to navigate in city environments \cite{Kayalibay2018NavigationMaps}. In order to infer long term dependencies and store pertinent information about the environment; network architectures typically incorporate recurrent memory such as Gated Recurrent Units \cite{ChungGatedNetworks} or Long Short-Term Memory \cite{Hochreiter1997LONGMEMORY}.\newline
The scaling of Deep RL has produced some impressive results, such as in the IMPALA \cite{impala2018} architecture which successfully trained an agent that can achieve human level performance in all 30 of the 3D, partially observable, DeepMind Lab \cite{Beattie2016DeepMindLab} tasks; accelerated methods such as in \cite{Stooke2018AcceleratedLearning} which solve many Atari environments in tens of minutes. The achievements in long term planing and strategy in DOTA \cite{OpenAI_dota} and StartCraft 2 \cite{vinyals2019grandmaster}, has demonstrated the Deep RL can learn policies with complex reasoning and long term decision making. There has been a recent push towards photo-realistic simulation, with the release of the Gibson \cite{xiazamirhe2018gibsonenv} dataset and the Habitat simulator \cite{habitat19iccv, szot2021habitat}. Many research teams focus on Structured Deep Reinforcement Learning, where priors are incorporated in the agent architecture, leading to advancements in GPS guided point-to-point navigation \cite{chaplot2020learning, chaplot2020neural}, memory-based tasks \cite{wayne2018unsupervised, beeching2020egomap, beeching2020learning}, semantic prediction \cite{chaplot2020object, chaplot2020semantic}. Other approaches aim to solve these challenging problems with scale, using distributed computing to training larger networks for billions of environment interactions \cite{wijmans2019dd, impala2018}. The video games industry has identified the advancements in Deep Reinforcement Learning as an opportunity to enrich video game experiences. With short term objectives being that of automated testing \cite{gordillo2021improving} and content generation \cite{gisslen2021adversarial}, and medium to longer term being player facing bots \cite{ijcai2021-294}, controlled by policies learned through interaction with the game environment. 
\begin{figure*}[t]
    \centering
    \includegraphics[width=0.90\textwidth]{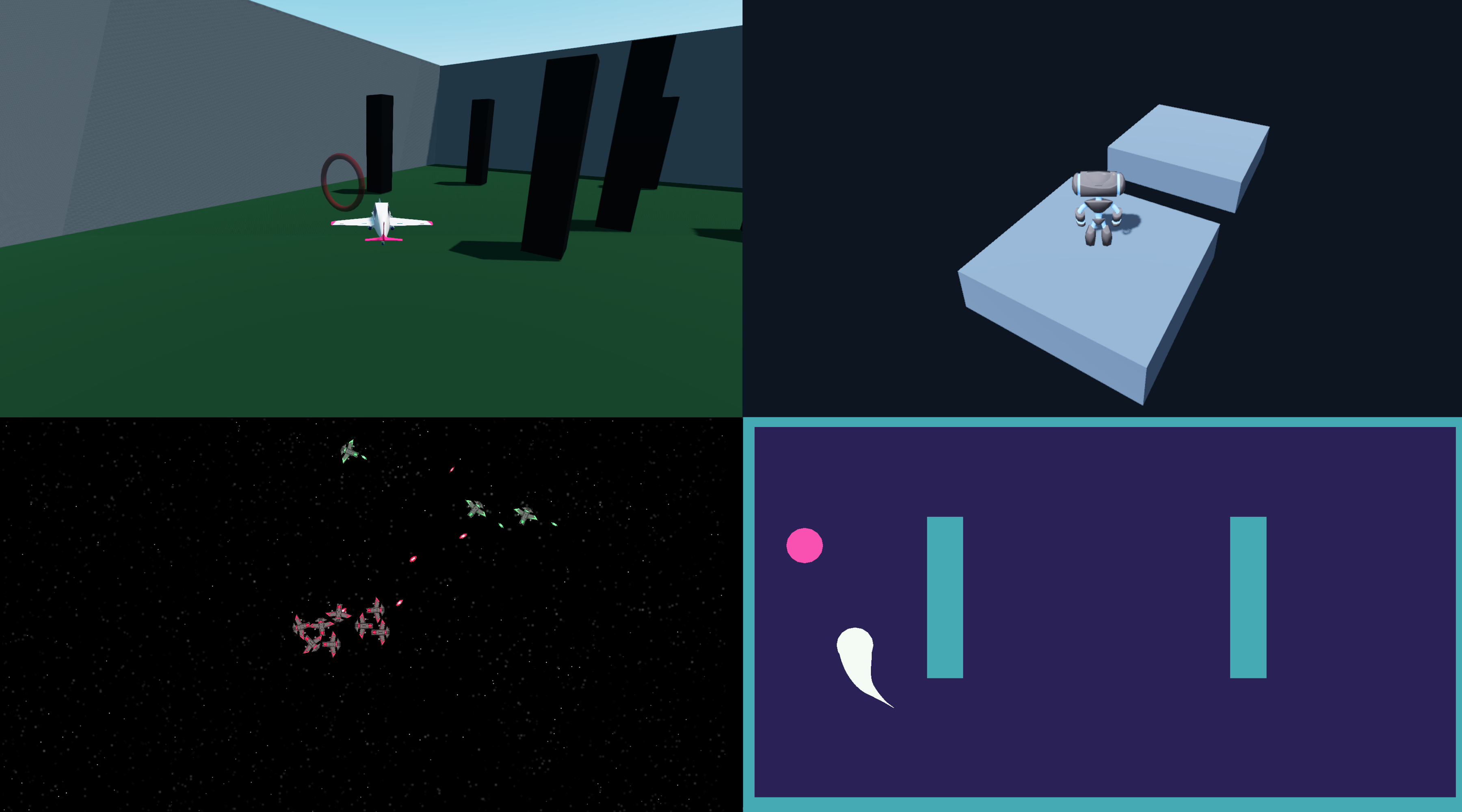}
    \caption{4 of the example environments available in the Godot RL Agents framework. Top left: \textit{Fly By}, where the agent must learn to fly a place between way-points in 3D space. Top right: \textit{Jumper}, where the objective is the navigated a bipedal robot between platforms, jumping where necessary. Bottom left: \textit{Space Shooter}, a 2D multi agent scenario, where one team must shoot and and destroy the agents from the other team. Bottom-right: \textit{Ball Chase}, where the agent must learn to collect pink fruits, while avoiding walls and obstacles.}
    \label{fig:example_envs}
\end{figure*}
\subsection{Deep RL environments}
There has been a large expansion to the number of Deep RL environments available in the last five years, the most well known being the gym framework \cite{Brockman2016OpenAIGym} which provides an interface to fully observable games such as the Atari-57 benchmark. Mujuco \cite{todorov2012mujoco} has remained the de facto standard for continuous control tasks, its recent open-source release can only benefit the larger RL community. OpenSpiel \cite{LanctotEtAl2019OpenSpiel} provides a framework for RL in games such as Chess, Poker, Go, TicTacToe and Habani, to name a few. There are a variety of partially observable 3D simulators available, with more game-like environments such as the ViZDoom simulator \cite{wydmuch2018vizdoom}, DeepMind-Lab \cite{Beattie2016DeepMindLab} and Malmo \cite{Johnson}. A number of photo-realistic simulators have also been released such as Gibson \cite{xiazamirhe2018gibsonenv}, AI2Thor \cite{KolveAI2-THOR:AI}, and the highly efficient Habitat-Lab simulator \cite{habitat19iccv,szot2021habitat}. The aforementioned environments were designed to train and evaluate RL agents in relatively small environments and typically are somewhat inflexible to addition of new tasks, abilities and action spaces. The Unity ML agents framework \cite{juliani2018unity}, provides flexibility in terms of design of environments and tasks, but with the downside of being a closed-source and pay to use (if your organization has a revenue of more than \$100k per year). In addition its closed-source implementation is inflexible if you wish to, for example, make changes to the protocol used to exchange observations between the environment and the RL algorithm. For these reasons the platform is platform somewhat prohibitive to the wider RL community.

\subsection{The Godot Game Engine}
The Godot Game Engine \cite{godot_game_engine} is an open-source tool for developing 2D and 3D games. Currently Godot is the most popular open-source Game Engine available and can export to MacOS, Linux, Windows, Android, iOS, HTML and web assembly. Godot supports multiple programming languages including Python, C\#, C++ and GDScript (a Python-like scripting language). The Game Engine provides an interactive editor for the design and creation of video game environments, characters, animations and menus, which enables fast iteration for prototyping new ideas. 
Godot has all the features of a modern Game Engine such physics simulation, custom animations, plugins and physically-based rendering.

\subsection{Deep RL Frameworks}
When it comes to applying a state of the art Deep RL algorithm, there are many quality open-source implementations available such as RLlib \cite{liang2018rllib}, ACME \cite{hoffman2020acme}, SampleFactory \cite{petrenko2020sf}, Seed RL \cite{espeholt2019seed}  and Stable Baselines \cite{stable-baselines} to name a few. Due to the availability of such high-quality, featured and well tested Deep RL frameworks; in this work we chose to design and implement an interface between the Godot Game Engine and two well known RL frameworks: RLlib \cite{liang2018rllib} and Stable Baselines \cite{stable-baselines}. Our objective is to focus on the interface and quality examples of 2D and 3D environments. This is contrary to the Unity ML Agents framework, which also implemented a limited set of RL algorithms.
\begin{figure*}[t]
    \centering
    \includegraphics[width=1.0\textwidth]{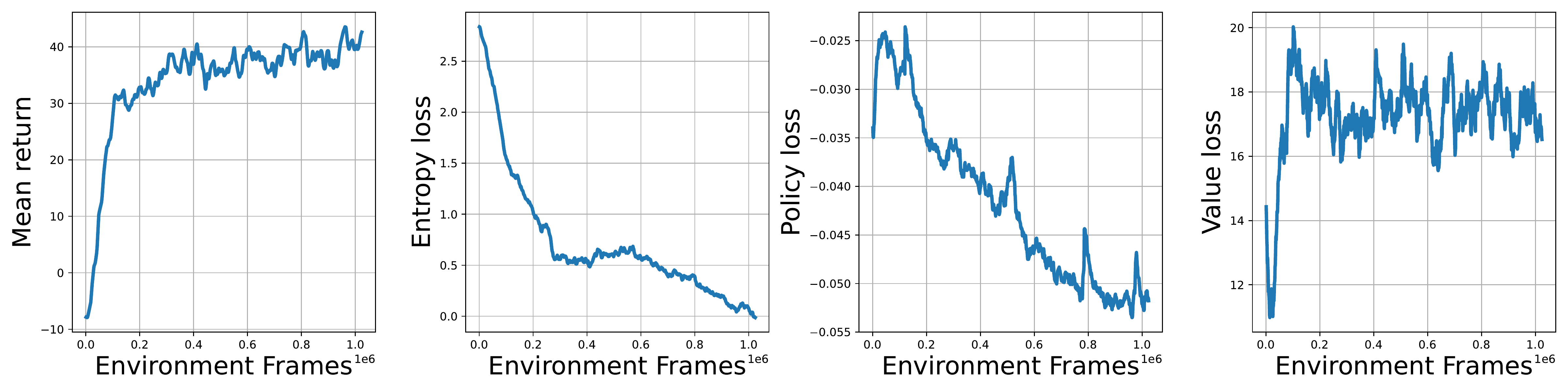}
    \caption{Training curves available in Godot RL Agents for the Ball Chase environment. Shown are a subset of available statistics output during training, Mean return, Entropy loss, Policy loss and Value loss}
    \label{fig:training_curves}
\end{figure*}
\section{Godot RL Agents}
The Godot RL Agents framework is an interface between the Godot Game Engine and Deep RL frameworks, allowing the creation of 2D, 3D and text-based environments, custom observations and reward functions. It is a flexible, fast and robust interface between RL training code running in Python and the Godot environment running interactively or as a compiled executable. We implement two forms of parallelization in the Godot RL Agents interface: in-game environment duplication and multiprocessing with several parallel executables. The interface has been designed to support continuous, discrete and mixed action spaces. Action repetition is implemented on the environment side to avoid unnecessary inter-process communication. Training can be performed interactively, while the environment is running in the Godot game editor: enabling mid-training pausing, analysis and debugging of variables. Once the user is confident with their environment configuration though interactive testing, the environment can be exported and run as an executable. Environment export enables accelerated, faster than real-time physics and headless training. The connection between the Game Engine and the Python training code is implemented with a TCP socket client-server relationship; this allows the environment to be run locally or distributed across many machines.

We provide a standard Gym wrapper and wrappers to Ray RLlib and Stable Baselines, which support the following algorithms.\\
On-policy algorithms:
\begin{itemize}
    \item A2C/A3C \cite{mnih2016asynchronous}
    \item PPO/APPO \cite{schulman2017proximal}
    \item IMPALA \cite{impala2018}
    \item DD-PPO \cite{wijmans2019dd}
\end{itemize}
Off-policy algorithms:
\begin{itemize}
    \item DQN \cite{mnih2013playing}
    \item Rainbow \cite{hessel2018rainbow}
    \item CQL \cite{kumar2020conservative}
    \item DDPG / TD3 \cite{lillicrap2015continuous}
    \item APEX \cite{horgan2018distributed}
    \item R2D2 \cite{kapturowski2018recurrent}
    \item SAC continuous \cite{haarnoja2018soft} and discrete \cite{christodoulou2019soft}.
\end{itemize}

Other algorithms are also available such as Behavior Cloning \cite{NEURIPS2018_4aec1b34}, Intrinsic curiosity \cite{pathak2017curiosity} and Evolutionary Strategies \cite{salimans2017evolution}. These pre-existing implementations enable researchers and game developers to rapidly evaluate different baselines architectures and algorithms, with robust implementations of state of the art RL algorithms. The detail the distributed architecture used in Godot RL Agents in Figure \ref{fig:learning_architecture}.

\subsection{Sensors}
Godot RL Agents allows users to add custom sensors to their agents to observe the environment. While simulated monocular cameras are possible, rendering is an expensive operation and rendered images do no always provide the more pertinent information. We provide implementations of circular and spherical "Raycast" sensors in both 2D and 3D that perform depth measurements, essentially a virtual LIDAR. Raycasts are a popular technique in video games and Game Engines such as Godot offer a highly optimized raycast implementation, making ray-casts a cheap option to augment an agent's observation. Other custom information, such as position, can be included in the agents observation, if the environment designer believes it is applicable to the task. 
\subsection{Example environments}
In version 0.1.0 of Godot RL Agents, we have implemented 4 example environments as a reference for researchers and game creators who wish to use our tool to build new worlds. We have created two 3D environments and two 2D environments. The source-code and further details (such as reward functions, reset conditions, action spaces) of these reference implementations, is available in our open-source repository \url{https://github.com/edbeeching/godot_rl_agents}. 

\subsection{Jumper}
Jumper is a 3D single agent environment where the agent must learn to jump from one randomly placed platform to the next. The action space is continuous for turning and movement, and discrete for the jump action. The episode is terminate if the agent falls off the platform. Observations are a cone of 3D ray-casts in front of the agent and vector pointing towards the next platform.

\subsection{Ball Chase}
Ball Chase is a 2D single agent environment where the agent must navigate around room and collect pink balls. The episode is terminated if the agent hits a wall. Observations are a circle of 2D ray-casts around the agent and vector pointing towards the next fruit. The action space is a continuous 2D vector indicating the move direction in x and y.

\subsection{Fly By}
Fly By is a 3D single environment where the agent must learn to fly from one way-point to the next. The episode is terminated if the agent hits a leave the game are. Observations are two vectors to the next two subsequent way-points.

\subsection{Space Shooter}
Space Shooter is a 2D multi-agent environment with two teams of 8 agents, where the objective is to shoot and destroy the enemy team. Currently observations are an array of relative positions of the enemy and friend teams. Results, observations and training methodology in the environment are still work in progress and will change in the next version of Godot RL Agents. 

\section{Experiments and Benchmarks}
As part of our reference implementation, we provide a PPO example with reasonable default parameters in order to provide a good starting point for training RL agents. We believe this algorithm is a suitable starting point, as it allows for both continuous and discrete action spaces, is robust to hyper-parameter configurations, and is relatively sample efficient. In Figure \ref{fig:training_curves} we show example training curves for the Ball Chase environment. We provide pre-trained models for all example environments in our repository, refer to our overview video for examples of these behaviors:

\noindent \url{https://youtu.be/g1MlZSFqIj4}.

\begin{figure}
    \centering
    \includegraphics[width=0.5\textwidth]{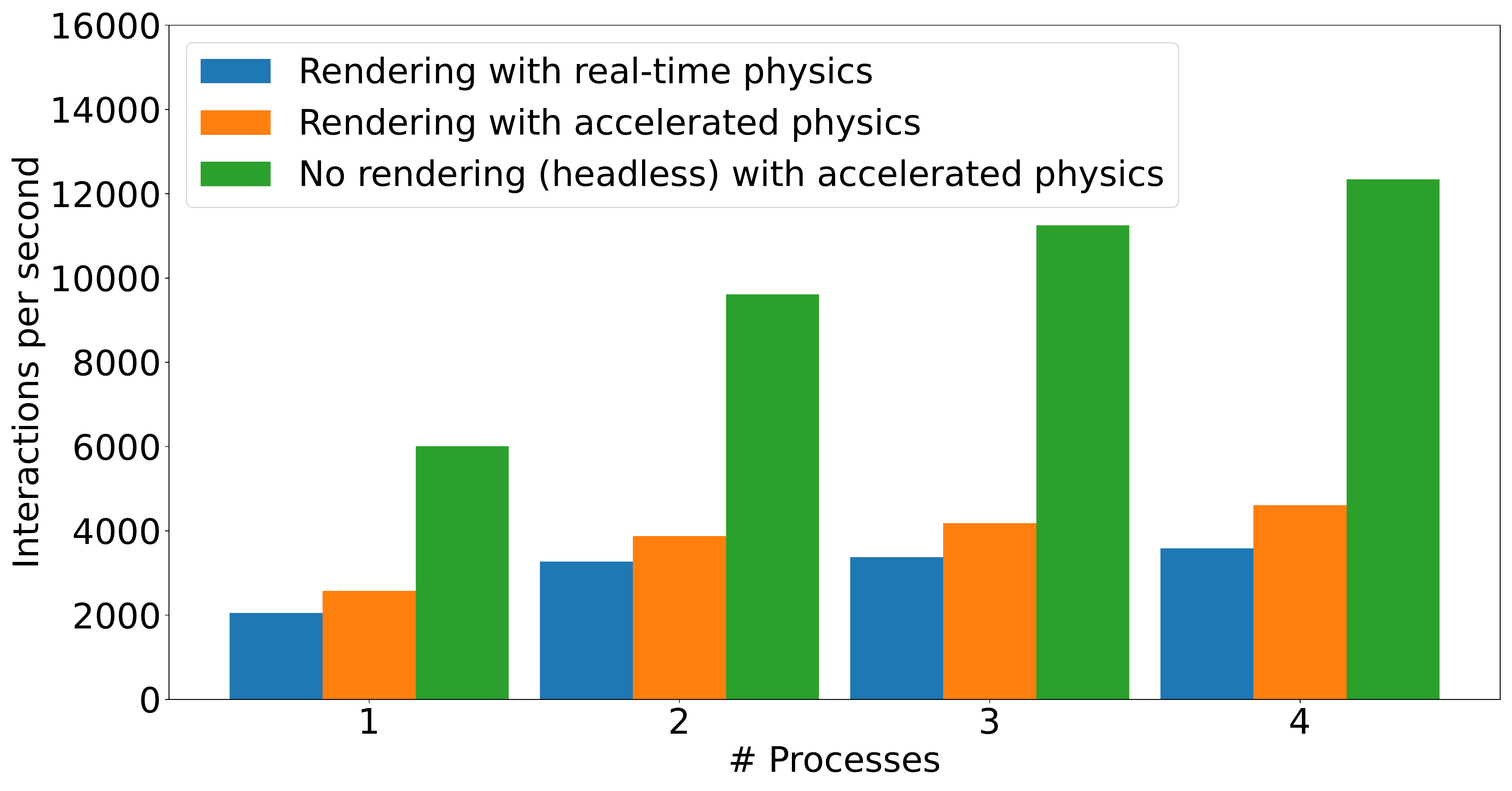}
    \caption{Benchmark of interactions per second when performing multi-process training in the Jumper environment with an action repeat of 4. We achieve 12k interactions per second with 4 parallel processes. Ideally scaling would be linear, future versions of Godot RL Agents will aim to profile and optimize the scaling of environment interactions.}
    \label{fig:benchmark}
\end{figure}

We benchmark the interaction rate of the Jumper environment and achieve 12k environment interactions per second with up to 4 parallel processes, results shown in Figure \ref{fig:benchmark}. Benchmarking was performed on a relatively high-end laptop computer \footnote{Dell XPS - i7-10750H (6 cores), GTX 1650 Ti, 32 GB RAM}, future work on Godot RL Agents will benchmark on dedicated hardware.
\section{Conclusions and future work}
We have introduced the Godot RL Agents framework, a tool for building 2D and 3D environments in the Godot Game Engine and learning agent behaviors with Deep Reinforcement Learning. We have created 4 example environments to demonstrate the capabilities of this versatile tool. By interfacing with well known open-source Deep RL implementations, we provide access to over 20 RL algorithms out of the box. We have implemented sensors,  tools for parallel interaction with environments, and support for continuous, discrete and mixed action spaces. 

Future work on the framework aims to include: variable size observations coupled with attention mechanisms, text-based environments, environments with discrete state spaces, virtual monocular cameras for vision-based agent control, multi-modal observations (1D, 2D, text \& audio) and model export for in-engine inference of trained behaviors. We plan to extend our focus to multi-agent environments, preliminary work in the \textit{Space Shooter} environment shows this area is a challenging area of research, with applications in both video games and collaborative robotics.

\begin{figure}
    \centering
    \includegraphics[width=0.5\textwidth]{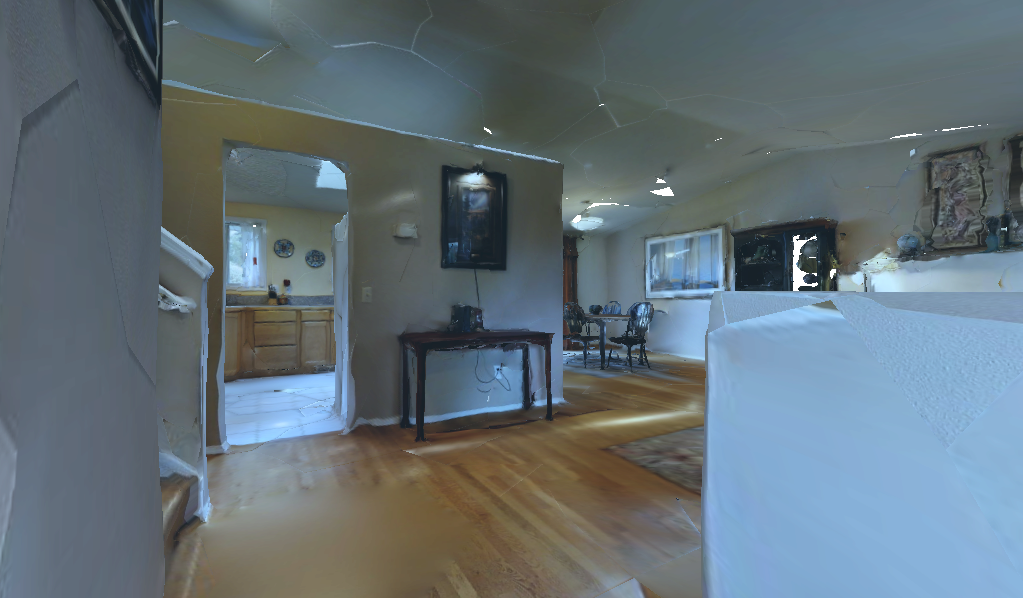}
    \caption{Future work, an example of a virtual monocular camera, rendering part of a scene from the Gibson dataset, loaded as an environment in the Godot RL Agents framework.}
    \label{fig:gibson_in_godot}
\end{figure}

External datasets can also be loaded in to the Godot Game Engine. In Figure \ref{fig:gibson_in_godot} we load photo-realistic scans from the Gibson \cite{xiazamirhe2018gibsonenv} dataset, to demonstrate that is tool can be used to build realistic 3D scenes for learning robotic control and sim2real transfer. We believe that there are near endless possibilities for this tool and are interesting to discuss with the research community the future direction of this work.

\bibliography{aaai22}

\end{document}